\pdfoutput=1
\documentclass[11pt]{article}

\usepackage{acl}

\usepackage{mathtools} % amsmath with fixes and additions
\usepackage{booktabs} % commands to create good-looking tables
\usepackage{tikz} % nice language for creating drawings and diagrams
\usepackage{lipsum}
\usepackage{xcolor}
\usepackage{amsmath}
\usepackage{booktabs}
\usepackage{amssymb}
\usepackage{mathtools}
\usepackage{amsfonts}
\usepackage{amsthm}
\usepackage{defs}
\usepackage{times}
\usepackage{latexsym}

\usepackage[T1]{fontenc}
\usepackage[utf8]{inputenc}

\usepackage{microtype}

\title{TransDrift: Modeling Word-Embedding Drift using Transformer}

\author{Nishtha Madaan, Prateek Chaudhury, Nishant Kumar, Srikanta Bedathur \\
Department of Computer Science\\
         Indian Institute of Technology Delhi India \\ anz208487@cse.iitd.ac.in, cs1190384@iitd.ac.in, cs5190586@iitd.ac.in, srikanta@cse.iitd.ac.in}

\begin{document}
\maketitle
\begin{abstract}
%Placeholder for now, have to fix it.
% Today we see the adoption of ML models in a plethora of downstream tasks and domains. For instance, automobile insurance providers leverage NLP models to validate claims, and recruiters also use ML models to analyze candidates resumes. But deploying models to automate business tasks is only half the job done. It is also critical to continuously monitor the deployed ML models, as once its deployed, we could encounter several issues in practice. In this work, we will limit our discussion to one of the issues previously discussed, i.e., where the performance of the deployed model degrades over time (aka drift). In practice, when the performance of the models degrades over time, it could have some severe consequences. For instance, in the cases of ML models deployed by an auto insurer, if they encounter texts from insurance claims not seen during the training of the ML model, the model could start approving invalid claims. Hence the deployments must be regularly monitored so that the stakeholders can be alerted when the model's performance or data properties change over time. If unaddressed, these issues could harm businesses as the corresponding stakeholders can no longer rely on the model's predictions.
% {\color{red}\lipsum[1]}
% {\color{red}\lipsum[1]}
In modern NLP applications, word embeddings are a crucial backbone that can be readily shared across a number of tasks. However as the text distributions change and word semantics evolve over time, the downstream applications using the embeddings can suffer if the word representations do not conform to the data drift. Thus, maintaining word embeddings to be consistent with the underlying data distribution is a key problem. In this work, we tackle this problem and propose TransDrift, a transformer-based prediction model for word embeddings. Leveraging the flexibility of transformer, our model accurately learns the dynamics of the embedding drift and predicts the future embedding. In experiments, we compare with existing methods and show that our model makes significantly more accurate predictions of the word embedding than the baselines. Crucially, by applying the predicted embeddings as a backbone for downstream classification tasks, we show that our embeddings lead to superior performance compared to the previous methods. We will release the code and datasets at \href{https://github.com/transdrift/} {\texttt{https://github.com/transdrift/}}
\end{abstract}

\section{Introduction}\label{sec:intro}
% {\color{red}\lipsum[1]}
\begin{figure}[t]
    \centering
    \includegraphics[width=\linewidth]{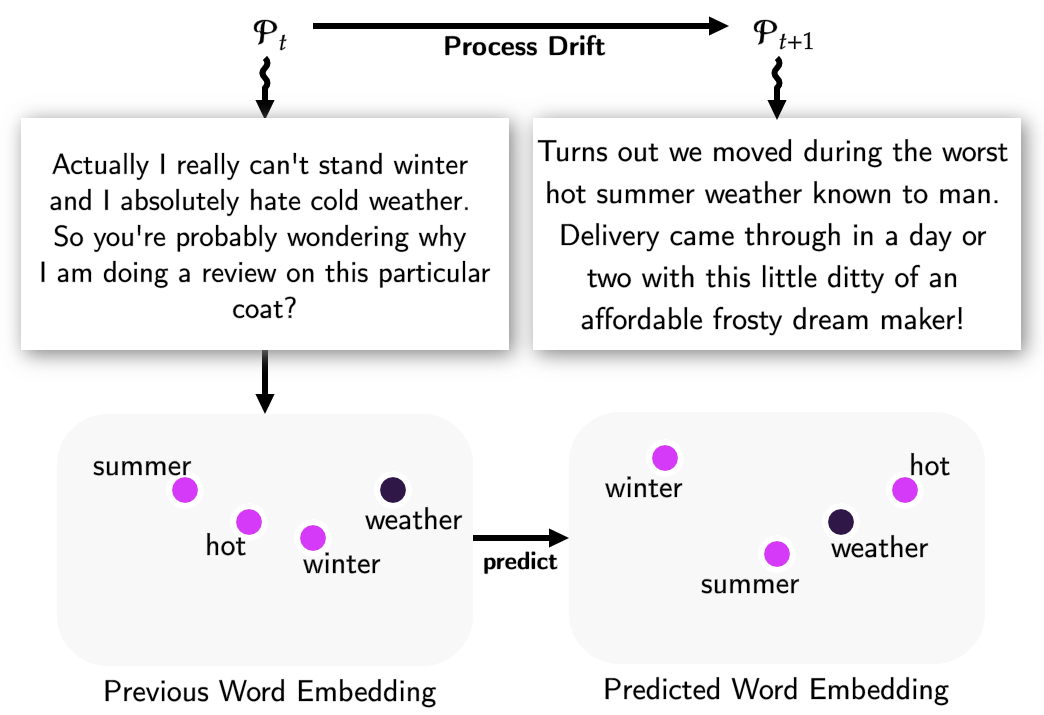}
    \caption{\textbf{Overview of Our Model.} We show an illustration of product review data showing that the data generating process undergoes drift between winter and summer. In these reviews, winters are characterized by the mentions of cold weather while summers are characterized by mentions of hot weather. Our model takes the past word embeddings as input to predict the embeddings for the drifted data distribution.}
    \label{fig:process_drift}
\end{figure}
% Para 1
Word embeddings play a major role in modern NLP applications providing a re-usable feature store that can be easily shared across multiple NLP tasks. This has led to their wide adoption in industry \citep{twitterfeature, tensorhub, feast, derczynski2015usfd, fromreide2014crowdsourcing}. A crucial aspect of the real world is that data distributions change over time. In text, words can gradually acquire new semantics and usage over time. For instance in the summer months, the word \textit{vacation} may be used more in relation to \textit{beach} than to \textit{skiing} and vice-versa in the winter months. A customer preference model relying on nearest neighbor lookup in the embedding space would thus suffer if the word embedding of \textit{vacation} is closer to \textit{beach} than to \textit{skiing} during winter. Hence, a good embedding needs to be consistent with such underlying changes in data distribution to be useful for downstream NLP applications. 

% \sbcomment{Does this continual learning also be relevant? see: \url{https://www.ijcai.org/Proceedings/2018/0627.pdf}, } % This approach requires access to past data and also requires some amount of data from the target domain (unlike ours).

In addressing this problem, a key concern is that when temporal drift occurs, there is not enough data available from the drifted distribution. For instance, when the summer ends and winter begins, we may have a large dataset collected during the summer months but very little data for the winter. When the model is deployed, a naive solution that we can simply re-train the embeddings from the winter data is not possible as the data is too small or sometimes does not even exist. However, we still seek an updated set of word embeddings that are consistent with the underlying data drift.

At the intersection of word embedding and data drift, previous works have analyzed historical data to identify temporal drifts in word embeddings \citep{hamilton2016cultural, hamilton2016diachronic,  dubossarsky2017outta,huang2018examining, kutuzov2018diachronic, garg2018word}. These works have shown not only that drift occurs but also identify the characteristics of the drift and their adverse effect on the performance of the downstream tasks. Instability of word embedding due to small data drifts has also been highlighted \citep{leszczynski2020understanding, Chugh2018StabilityOW, hellrich2016bad, antoniak-mimno-2018-evaluating}. However these approaches do not provide a way to learn the dynamics of drift for updating the word embeddings consistently with the data drift.

% learn the dynamics of the drift and predictively infer the word embeddings from sparse or no data. 

% combines the knowledge from the word embedding of the past month with the knowledge of the drift dynamics.

% and predicts good word embedding for future distribution without requiring large training datasets.

In this work, we propose TransDrift, a novel model that learns to predict the future word embeddings that are consistent with the drift in data. Our model combines the knowledge from the past word embedding with the knowledge of the drift dynamics and predicts accurate future word embedding. Our solution utilizes the flexibility of the transformer architecture to make the predictions. Our experiments demonstrate the effectiveness of using the predicted embeddings as backbone for downstream NLP tasks. Crucially, our method is simple and general can be used along with any word embedding algorithm.

Our main contributions can be summarized as: \textit{1)} We propose the first model, TransDrift, that leverages transformer to predict the future embeddings. \textit{2)} Our model can predict future embeddings. \textit{3)} Our results show that our model is effective in modeling the drift in the embeddings. \textit{4)} Lastly, we also show improvement in the accuracy on downstream NLP tasks when using our predicted embeddings.

% TransDrift -- 1) first system, 2) flexible due to transformers, 3) can learn from sparse data, 4) helps improving downstream model accuracy, 5) Embedding technique is not specific to embedding generation algorithm. 

% In NLP applications the text distribution can change rapidly with time. 
% Using such time-aware word embeddings, the downstream NLP system can make more accurate inferences.
% By predicting the word-embedding, we can use them as a backbone for a downstream model instead of re-training the embedding with large amounts of data or having to re-use the old embedding.

% Para 2

% The text distribution can be called $\cP_t$.

% giving rise to data sample $\cD_t \sim \cP_t$

% And this undergoes temporal drift $\cP_1 \rightarrow \cP_2 \rightarrow \ldots \rightarrow \cP_T$

% The samples of data that we get become $\cD_1 \rightarrow \cD_2 \rightarrow \ldots \rightarrow \cD_T$
% , as a result, whose characteristics also change with time.

% For instance, only a few days have passed and not much data has been collected for the second month.

\section{Background}

\subsection{Word Embedding}

% Word embeddings provide a re-usable feature store that can be easily shared across multiple NLP applications which has led to their wide adoption in industry \citep{twitterfeature, tensorhub, feast}.
Several commonly used methods such as \textit{word2vec} generate rich application-agnostic embeddings for the words in the vocabulary \citep{word2vec2013, bojanowski2017fastText, pennington2014glove}. 
This process takes a text corpus $\cD$ and returns embeddings $\bE = \{\bee_1,\ldots,\bee_N$\} for $N$ words in the vocabulary. Here, $\bee_n$ represents $n$-th word in the vocabulary and it is a $d$-dimensional vector.  In these methods, the main goal is to embed the words in a feature space while capturing their underlying semantic structure. For instance, the words having a similar usage such as \textit{apple} and \textit{orange} are embedded close together in the feature space. To learn such embedding, the common approach is to take each word in the given text corpus $D$ and predict which words are present in its neighborhood. From this prediction objective, the gradient is backpropagated to the embedding of the input word which leads to learning of the word embeddings. Hence, the information about the neighborhood in which the word is commonly used becomes encoded in these embeddings.

\subsection{Transformer}
% {\color{red}\lipsum[3]}
Transformer is an architecture for processing a set of vectors such that each vector is updated by flexibly interacting with all the other input vectors \citep{transformers, settransformer}. Formally, given a set of $N$ vectors, a \textit{transformer layer} maps the input vectors to $N$ output vectors. To enable interaction among the input vectors, a transformer layer first performs self-attention between the vectors \citep{transformers}. Following this self-attention step, each vector is then fed to an MLP to generate the output vectors which makes the model more expressive. Residual connections are added to both the self-attention and the MLP steps for improved gradient flow. In practice, multiple transformer blocks are stacked together to increase the modeling capacity of the Transformer. As transformers have been shown to be a powerful architecture showing impressive performance by modeling complex interactions, in this paper we seek to bring this idea to track the drift in the word embeddings over time. 
% In this paper, we use transformers to track the drift in embeddings by training the model on embeddings from timestep $t$ and predicting the embeddings at timestep $t+1$. 
% e_1, e_2, \ldots e_T

\section{Method}

% \subsection{Temporal Drift and Data Sparsity}

In this section, we propose a simple method to model the drift in word embeddings over time. Consider the text distribution at each time-step $\cP_t$ which provides us a data sample $\cD_t \sim \cP_t$. Crucially, this distribution undergoes change over time: $\cP_t \rightarrow \cP_{t+1}$. Resulting from this distribution change, the word semantics and usage in the sampled datasets, $\cD_t$ and $\cD_{t+1}$, also change with time. We would like these changing semantics to be reflected in word embeddings of each time-step for them to be useful for the downstream tasks. That is, we desire word embeddings $\bE_t$ at each time-step $t$ such that they are consistent with the changing data distribution.

A key consideration is that while data at time $t$ is large, the data $\cD_{t+1}$ at time $t+1$ is significantly smaller and might even be an empty set. Thus, while word embedding $\bE_t$ can be learned accurately from $\cD_t$ using standard methods such as \textit{word2vec}, however, directly learning $\bE_{t+1}$ from $\cD_{t+1}$ is likely to be ineffective and even impossible if $\cD_{t+1}$ is empty. Hence, during training, we seek to learn the drift dynamics that can be utilized at test time to predict $\bE_{t+1}$ directly from $\bE_{t}$ even when the data $\cD_{t+1}$ from the time-step $t+1$ is small or empty.

For this, we first train the embedding at time $t$ using the large data set at time $t$ and then use a Transformer to map the word embeddings at time-step $t$ to the embeddings of the next time-step $t+1$. Formally, 
\begin{align}
    {\bE}_{t} &= \text{TrainWordEmbeddings}(\cD_t),\nn\\
    {\bE}_{t+1} &= \text{Transformer}_\phi({\bE}_{t}).\nn
\end{align}
By using a Transformer model, the prediction of each word embedding can see the embeddings of all the other words via attention. This allows our model to learn complex dynamics of embedding drift, helping the model make better predictions.

Optionally, at inference time, if we have a small amount of data $\cD_{t+1}^\text{small}$ at timestep $t+1$, we can use it to train the embeddings of a small number of words. The resulting embeddings ${\bE}_{t+1}^\text{small}$ can be provided as an additional context to our model during prediction. Taking these in addition to the embeddings of the previous time-step, our model predicts all the embeddings of time-step $t+1$. This can be summarized as follows: 
\begin{align}
    {\bE}_{t+1}^\text{small} &= \text{TrainWordEmbeddings}(\cD_{t+1}^\text{small}),\nn\\
    {\bE}_{t+1} &= \text{Transformer}_\phi({\bE}_{t}, {\bE}_{t+1}^\text{small}).\nn
\end{align}
% where ${\bE}_{t+1}^\cC$ are optional context embedding on any available dataset of that time-step which, we shall show, 
In our experiments, we shall show that providing such additional context can lead to moderate improvements in the prediction accuracy. For downstream applications, providing such additional embeddings can therefore be beneficial.

\begin{figure}
    \centering
    \includegraphics[width=\linewidth]{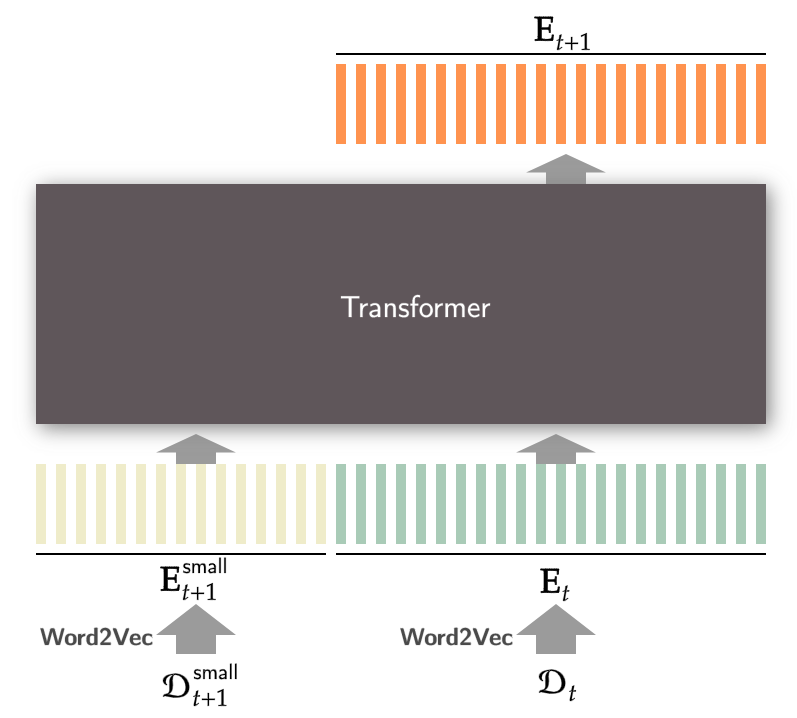}
    \caption{\textbf{Our Model.} Our model takes the word embedding of previous time-step as input optionally along with a few word embeddings of the current time-step trained with a small dataset.}
    \label{fig:arch}
\end{figure}

\textbf{Training.} For training, we assume that our historical data provides large datasets for both time-steps $t$ and $t+1$ which we denote as $\cD_t$ and $\cD_{t+1}$. Taking these two datasets, we train the word embeddings as follows:
\begin{align}
    {\bE}_{t} &= \text{TrainWordEmbeddings}(\cD_{t}),\nn\\
     {\bE}_{t+1} &= \text{TrainWordEmbeddings}(\cD_{t+1}).\nn
\end{align}
To train the Transformer, we minimize the following cosine embedding loss $\cL^\text{predict}(\phi)$ for predicting the embedding at time $t+1$:
% \begin{align}
%     \cL^\text{tracker}(\phi) = \| {\bE}_{t+1}  - \text{Transformer}_\phi({\bE}_{t}, {\bE}_{t+1}^\text{small}) \|^2.\nn
% \end{align}
\begin{align*}
    1 - &\text{cos}(\bE_{t+1}, \text{Transformer}_\phi({\bE}_{t}, {\bE}_{t+1}^\text{small})),\nn
\end{align*}
where $\text{cos}(\cdot, \cdot)$ denotes cosine similarity.

\textbf{Downstream Task.} As our end goal of modeling the embedding drift is to help downstream task, we now describe how we utilize our predicted word embedding to achieve this. We train a downstream task neural network on the predicted word embeddings as follows:
\begin{align}
    f_\ta(\bx; \bE_{t+1}).\nn
\end{align}
Given embedding $\bE_{t+1}$, input $\bx$ and target label $\by$, we learn the task-specific neural network at time-step $t+1$ as:
\begin{align}
    \mathcal{L}^\text{task}(\ta) = \text{CrossEntropy}(\by, f_\ta(\bx; \bE_{t+1})).\nn
\end{align}

% {\color{red}\lipsum[3]}

% {\color{red}\lipsum[3]}

% {\color{red}\lipsum[3]}

% {\color{red}\lipsum[3]}
 
% {\color{red}\lipsum[3]}

% Lifelong Domain Word Embedding via Meta-Learning 
% requires access to the past corpora.
% cannot work with no data of the drifted domain.

% In modeling the drift in word embedding, 

% -----  Limited expressiveness: Assume same shift for all words applied as a linear function of time. Seems rather restrictive. Unsound method: Applying sentence embedding drift to word embedding seems not reasonable. Motivation seems odd:  Adding embedding shift in reverse is not good, because 'holiday' may mean 'surfing' in summer but 'skiing' in winter. By doing a reverse, we are essentially saying that in winter, 'skiing' is not that important but surfing is -- which is clearly not desired. Also, then why do a reverse when you can simply freeze the embedding from the source dataset.   -----

% ----- \cite{hamilton2016diachronic} ------

%  -- evolutionary chains

% \textbf{Stability of Word Embedding.}

% --  ----

% --------

% --  --- 

% ------- ------

% -- These works do not model the drift over time. All these works assume the same data distribution while ours assumes the data is being drifted. These only identify and mitigate some instability issues when the data is same. --

% \citep{yang2019adaptive}

% \citep{Chugh2018StabilityOW}
% ----  \citep{pierrejean2018predicting} -----

% ---- \citep{antoniak-mimno-2018-evaluating} ---

% ---- \citep{huang2020measuring}----

% {\color{red}\lipsum[3]}

% {\color{red}\lipsum[3]}

% {\color{red}\lipsum[3]}

% {\color{red}\lipsum[3]}

\section{Experiments}
The goal of our experiments is to show how well our model can accurately predict the drifted word embeddings relying on little to no data from the drifted text distribution. Furthermore, we also show the benefits of our predicted word embedding in improving the performance of downstream classification tasks. As an instance to test our idea, we intentionally choose simplest and widely used embedding method word2vec so that our results can be interpreted more generally.

\subsection{Experiment Setup}

\subsubsection{Datasets}
We evaluate our models on a synthetic dataset, Yelp Academic dataset \citep{yelpdataset} and Amazon Customer Review dataset \citep{amazonreviewdataset}. For each dataset, we consider drift instances with each instance consisting of $\cD_1$, $\cD_2$, and $\cD_2^\text{small}$ during training. Here, the subscript 1 denotes the source time-step $t=1$ and subscript 2 denotes the next time-step (i.e. $t=2$) in which the underlying text distribution has undergone a shift. 
%Out of these 1000 instances, we use 800 for training, 100 for validation, and the rest for testing.

\textbf{Synthetic Dataset.} This datasets consists of several instances. For each instance, we generate a random sparse graph with each node in the graph representing a token in the vocabulary. In this graph, we randomly assign edge weights to denote the co-occurence pattern among the tokens. Starting from a random node, we perform a random walk on the graph and collect the tokens encountered as text. During the random walk, the probability of transitioning to the next node is proportional to its edge weight. This results in a sample of dataset $\cD_1$. We then apply random modifications to edge weights of the graph. This modified graph is considered as the drifted data generating process. From this drifted graph, we again perform a random walk to sample a small dataset $\cD_2^\text{small}$ and a large dataset $\cD_2$.

\textbf{Yelp Academic Dataset.} For Yelp Academic Dataset, we use the businesses, reviews, and user data. For this, we divide the dataset into two parts by timestamp -- reviews before the year 2016 and reviews after the year 2016. We denote these two parts as: $\cD_1$ and $\cD_2$. We take smaller subsets of $\cD_2$ to obtain $\cD_2^\text{small}$.

\textbf{Amazon Customer Review Dataset.} For Amazon Customer Review dataset, we separately consider the categories: \textit{Books}, \textit{Electronics}, \textit{DVD}, and \textit{Kitchen}. For this, we divide the dataset into two parts by timestamp -- summer reviews and winter reviews. We call these two parts as: $\cD_1$ and $\cD_2$. We take smaller subsets of $\cD_2$ to be $\cD_2^\text{small}$. For all the datasets we discussed above, more details can be found in Appendix \ref{sec:appendix}.

% Next, $\bE_1$, $\bE_2$ and $\bE_2^\text{small}$ are obtained from these to form one sample input.

% {Each input sample comprises $\bE_1$, $\bE_2$, and $\bE_2^\text{small}$, representing correspondingly the Word2vec embeddings of $\cD_1$, $\cD_2$, and $\cD_2^\text{small}$. 

% Amazon Review also used in \citep{}
% {}

\subsubsection{Metrics}
To evaluate how well our predictions match the desired word embeddings of the drifted distribution, we compute cosine similarity between $\bE_2$ learned using full dataset $\cD_2$ and our predicted embeddings generated using the previous embeddings $\bE_1$ and $\bE_2^\text{small}$. To measure the benefits of our predicted embedding on downstream task, we report the accuracy of the predictions of the downstream model. 

% the similarity in sentiment prediction (on a per review basis) by $\bE_1$ and predicted embedding with respect to $\bE_2$.
% We compare the quality of embeddings $\bE_1$, $\bE_2$, and predicted $\bE_2$ by their sentiment prediction accuracy on the test dataset.

\subsubsection{Baselines}
As no previous work directly tackles our problem setting, we develop the following baselines to show the efficacy of our model.

\textbf{No-Drift Model.} In this baseline for predicting the future embeddings, the modeling assumption is that the word embeddings do not undergo drift. That is, the model assumes that the embeddings learned at time-step 1 using $\cD_1$ can be naively re-used time-step 2 even though the underlying data distribution has drifted between timesteps $1$ and $2$. The goal of this comparison is to justify the need for predicting the word embedding instead of simply re-using the previous outdated embeddings.

\textbf{Additive-Drift Model.} In this baseline for modeling the embedding drift, we assume that the drift can be modeled by adding a constant embedding vector to all the words in vocabulary as proposed by \cite{stowe2021combating}. That is, this model learns a vector $\Delta$ such that the embedding at time-step $2$ can be predicted as $\bE_2 = \bE_1 + \Delta$. The goal of this comparison is to show that it is not enough to simply model the drift as a constant additive vector and it is required to model complex interaction and non-linear drift dynamics to predict the future embedding accurately.

% \textbf{Baselines.}  {
% We calculate $\mu_1$ and $\mu_2$ by averaging vectors of all $\bE_1$s and $\bE_2$s, respectively. For predicting $\bE_2$ in the test dataset, add ($\mu_2-\mu_1$)  to every word of $\bE_1$. Later, calculate the average cosine similarity between predicted $\bE_2$ and $\bE_2$ to get the accuracy.
% }

% \textbf{Drift. } 
% Randomly selected some edges of some of the nodes selected randomly from the graph. Changed the weights of these edges by multiplying them by a random decimal between (0,2). This technique is used to introduce drift into synthetic datasets.
% \subsection{Synthetic Dataset}

% % \textbf{Embedding Tracking.} 
% % {For the Synthetic dataset, we consider a vocabulary of 100 words, manually selected, for our experiments. We implement a random sparse graph. Each node in the graph represents a word in the vocabulary, and the edge weights incorporate the word's context in correspondence to its neighbors. We now perform a random walk on the graph to obtain $\cD_1$. Sample a drift\textbf{*} in the graph and again perform a random walk on the drifted graph to get $\cD_2^\text{small}$ and $\cD_2$. Next, $\bE_1$, $\bE_2$ and $\bE_2^\text{small}$} are obtained from these to form one sample input.

% {We run the transformer model on the synthetic data-sets; $\bE_1$ and $\bE_2^\text{small}$ are given as input and $\bE_2$ as the label. Model is trained on 800 input samples and predicts the $\bE_2$ for the 100 test samples. We repeat this experiment by changing the vocabulary size of $\cD_2^\text{small}$, correspondingly changing $\bE_2^\text{small}$.
% }

\begin{table}[t]
\centering
\scalebox{0.85}{
\begin{tabular}{lllc}
\toprule
\textbf{} & \multicolumn{3}{c}{Drift Model}                \\ \cmidrule(l){2-4} 
Dataset   & No-Drift & Additive &  TransDrift      \\ \midrule
Synthetic & 0.32     & 0.33     &   \textbf{0.7724} \\
Yelp      & 0.19     & 0.7956   &   \textbf{0.8910} \\
Amazon    & -0.004   & -0.0002  &  \textbf{0.8170} \\ 
\bottomrule
\end{tabular}
}
% \end{table}

% \begin{table}[t]
% \centering
% \begin{tabular}{@{}lccc@{}}
% \toprule
% & \multicolumn{3}{c}{Drift Model} \\\cmidrule(l){2-4}
%   Dataset            & {No-Drift} & {Additive} &
%   {TransDrift} \\ \midrule
% Synthetic            & 0.32     & 0.33      &     \textbf{0.7724}          \\ 
% Yelp & 0.19 & 0.7956  &  \textbf{0.8910} \\ 
% Amazon             & -0.004   & -0.0002          &  \textbf{0.8170}             \\ \bottomrule
% \end{tabular}
\caption{\textbf{Comparison of word embedding prediction between our model and the baselines.} We report the cosine similarity of the predicted embedding with the ground truth embedding trained using large amount of data from the drifted distribution. The predicted embeddings do not use any data from the drifted distribution. We note that our model, TransDrift, is significantly more accurate with respect to the baseline models.}
\label{tab:quant-cosine-baselines}
\end{table}

\begin{table}[t]
\centering
\begin{tabular}{@{}lccccc@{}}
\toprule
% \multicolumn{1}{l}{} & \multicolumn{5}{c}{Cosine Similarity between P and E2}                     \\ \midrule
& \multicolumn{3}{c}{Size of $\cD_2^\text{small}$ as $\%$ of $\cD_2$ }\\ \cmidrule(l){2-4}
{Dataset} & 30\% & 20\% & 0\% \\ \midrule
Synthetic               & \textbf{0.8067}        & 0.7913        & 0.7724       \\ 
Yelp & \textbf{0.9119} & 0.9075 &  0.8910 \\ 
Amazon           & \textbf{0.8829}        & 0.8076        & 0.8170
\\ \bottomrule
\end{tabular}
\caption{Comparison of the word embedding prediction performance under varying percentages of $\cD_2^\text{small}$ data used. We report the average cosine similarity.}
\label{tab:quant-d2-variation}
\end{table}

\begin{figure*}[t]
    \centering
    \includegraphics[width=0.85\linewidth]{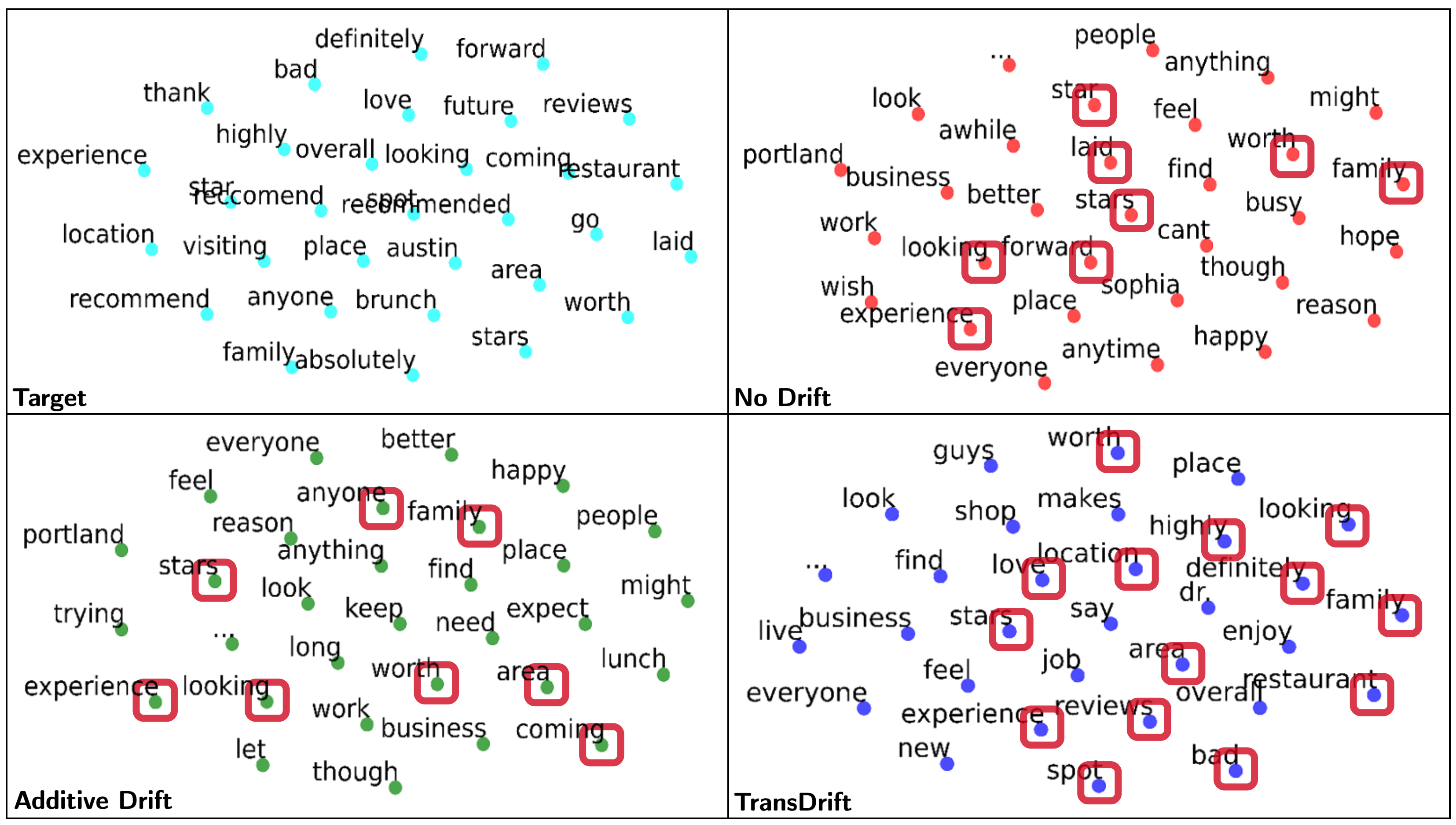}
    \caption{\textbf{Qualitative Comparison of TransDrift with the Baselines on Yelp.} We show the nearest neighbors of the word \textit{place} using the word embedding predictions of various models. We visualize the embeddings on a 2D plane using t-SNE. For each model, we highlight the nearest neighbors that match the target nearest neighbors (top-left) using red boxes. We see that our model, TransDrift, has the most number of common nearest neighbors with respect to the target (bottom-right). In contrast, the baselines, No-Drift and Additive Drift, have significantly fewer common nearest neighbors.}
    \label{fig:qualitative-1}
\end{figure*}

\begin{figure*}[]
    \centering
    \includegraphics[width=0.85\linewidth]{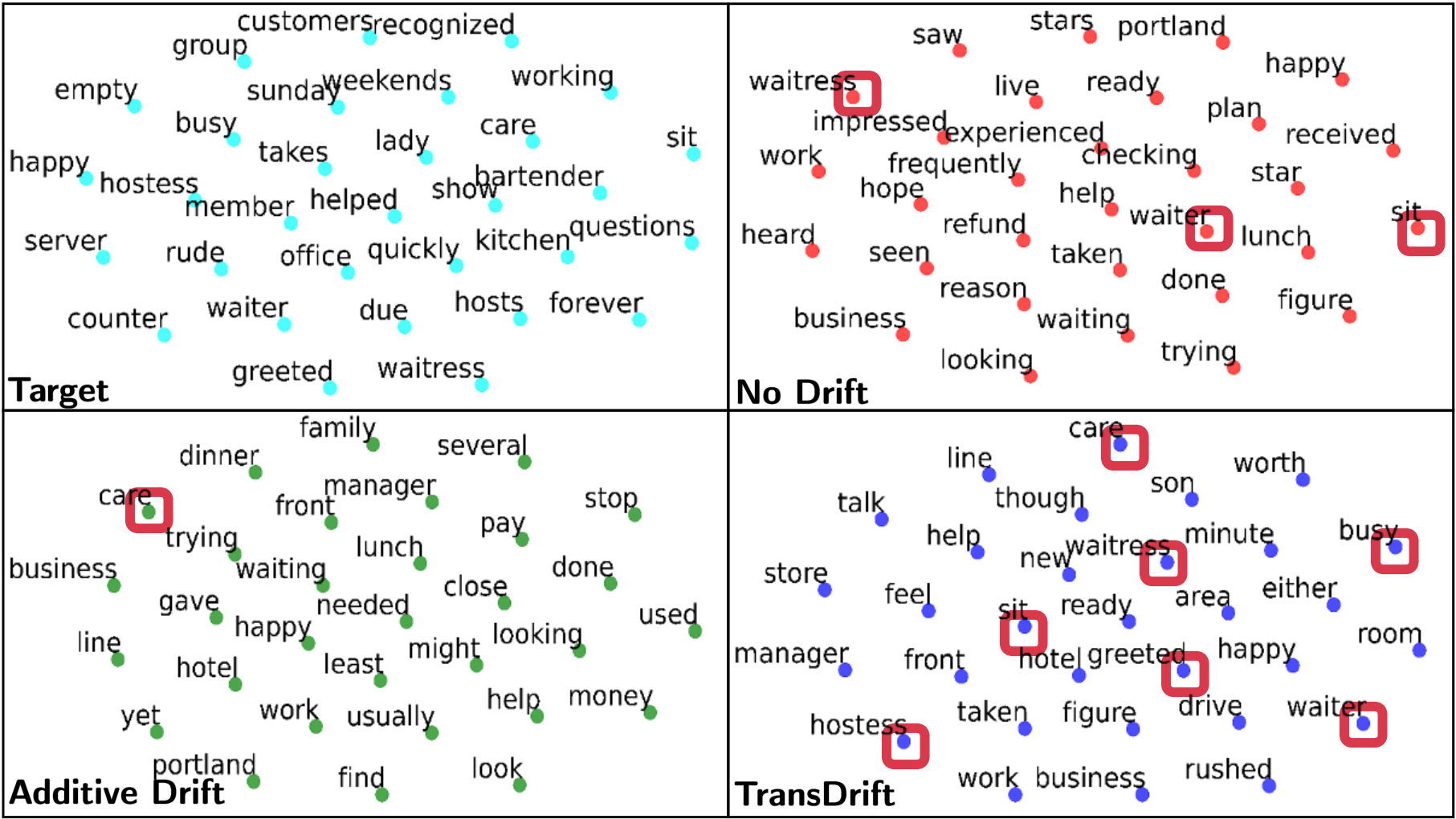}
    
    \vspace{5mm}
    
    \includegraphics[width=0.85\linewidth]{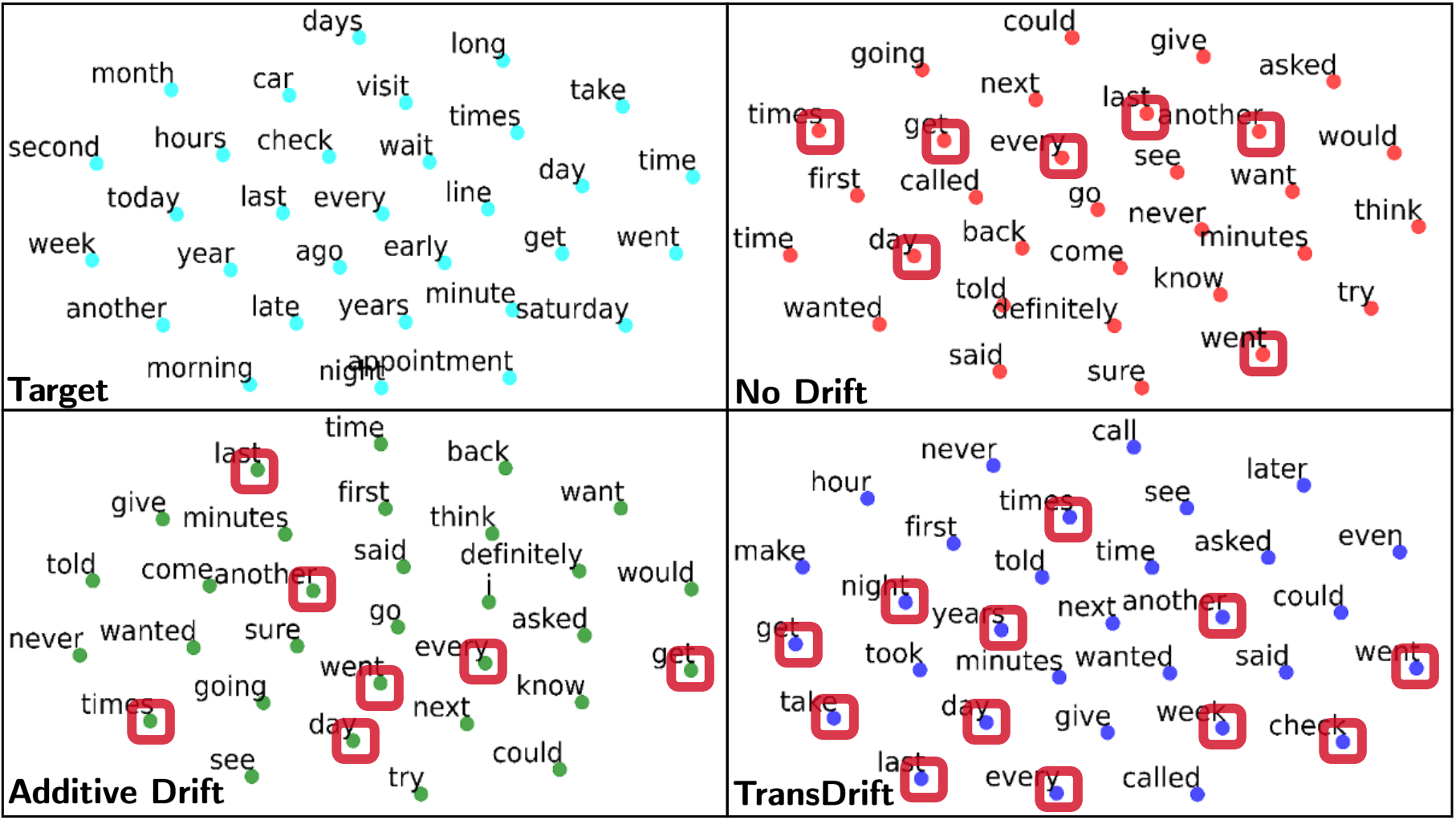}
    \caption{\textbf{Qualitative Comparison of TransDrift with the Baselines on Yelp.} We show the nearest neighbors of the word \textit{happy} (top) and \textit{time} (bottom) using the word embedding predictions of various models. We visualize the embeddings on a 2D plane using t-SNE. For each model, we highlight the nearest neighbors that match the target nearest neighbors (top-left) using red boxes. We see that our model, TransDrift, has the most number of common nearest neighbors with respect to the target (bottom-right). In contrast, the baselines, No-Drift and Additive Drift, have significantly fewer common nearest neighbors.}
    \label{fig:qualitative-2}
\end{figure*}

% \begin{figure*}[]
%     \centering
%     \includegraphics[width=0.9\linewidth]{images/time.png}
%     \caption{\textbf{Qualitative Comparison of TransDrift with the Baselines on Yelp.} We show the nearest neighbors of the word \textit{time} using the word embedding predictions of various models. We visualize the embeddings on a 2D plane using t-SNE. For each model, we highlight the nearest neighbors that match the target nearest neighbors (top-left) using red boxes. We see that our model, TransDrift, has the most number of common nearest neighbors with respect to the target (bottom-right). In contrast, the baselines, No-Drift and Additive Drift, have significantly fewer common nearest neighbors.}
%     \label{fig:qualitative}
% \end{figure*}

\subsection{Word-Embedding Prediction}
We now evaluate the performance of word embedding prediction by the models.

\subsubsection{Quantitative Evaluation}
We perform a quantitative evaluation by reporting the average cosine similarity under two prediction regimes: with and without the available data from the drifted distribution.

\textbf{Prediction with No Data.} In Table \ref{tab:quant-cosine-baselines}, we report the average cosine similarity of the predicted word embeddings with the target embeddings. We take the target embeddings $\bE_2$ to be those trained by applying \textit{word2vec} on a large dataset collected from the drifted distribution. In this comparison, we use no data from the drifted distribution for making the embedding predictions.  We show that our model TransDrift outperforms all the baselines significantly across all datasets. In particular, we see a poor performance of the No-Drift baseline which shows that it is not enough to simply re-use the outdated embeddings. Furthermore, when we assume that drift is a constant vector added to all the words (i.e. the Additive-Drift model), we see that the performance is better than the No-Drift model but is still significantly worse than our model. This shows that modeling complex interactions and non-linear drift behavior is crucial and is successfully modeled by our transformer-based predictor. 

% From table 1, the accuracy shown by the embeddings predicted by the transformer is much higher than the baseline accuracy, discussed above. 
% The no-drift column represents the cosine similarity within $\bE_1$ and $\bE_2$ showing the case where we are basically not taking drift into account. 
% Again these values are much lower than the accuracy given by embeddings predicted by the transformer. 

\textbf{Prediction with Available Drifted Data.} In Table \ref{tab:quant-d2-variation}, we show the effect of using increasingly larger amount of data $\cD_2^\text{small}$ from time-step 2 to inform the word embedding prediction in our model. We note that with increasing the size of this data, we see an increase in prediction accuracy across all datasets. In deployment settings, this property may be useful to continually improve the embeddings as increasingly more data is gradually collected. Interestingly, we note that even with no data from the time-step 2, our prediction accuracy already surpasses all our baselines reported in Table \ref{tab:quant-cosine-baselines} across all datasets. 

% Now, from table 3, we expect the accuracy for the prediction to increase along with the increase in the amount of $\cD_2$ present in $\cD_2^\text{small}$. 
% We perform the experiments by varying the size of $\cD_2^\text{small}$ vocab as 0, 20 and 30. As evidenced by the rise in accuracy form 0.7724 to 0.8067, there is a considerable improvement in embedding prediction with an increase in the size of the $\cD_2^\text{small}$ vocab.

% \subsubsection{Qualitative Results}
% \usepackage{booktabs}
% \usepackage{booktabs}
% \usepackage{booktabs}
% \usepackage{booktabs}

\subsubsection{Qualitative Evaluation}
To analyze our prediction results qualitatively, we take eight words: \textit{well}, \textit{place}, \textit{great}, \textit{time}, \textit{nice}, \textit{customer}, \textit{happy} and \textit{people} and visualize their nearest neighbors using the predicted embeddings of all the models. We also consider the word embedding trained using a large amount of data from time-step 2 to be the target embeddings. Hence, if a model is effective, then the the number of words common nearest neighbors between the predicted and the target embeddings would be larger. We visualize these in Figures \ref{fig:qualitative-1} and \ref{fig:qualitative-2}. We also report the number of common nearest neighbors for each word in Table \ref{tab:qual-nn-count}. We see that assuming No-Drift leads to few nearest neighbors suggesting that the underlying data drift indeed changes the word usage over time. However, when we use TransDrift to predict the embeddings, we see the most number of nearest neighbors as compared to the No-Drift and the Additive Drift baselines. 
\begin{table}[t]
\centering
\begin{tabular}{@{}lccc@{}}
\toprule
& \multicolumn{3}{c}{\# Common Neighbors}\\\cmidrule(l){2-4}
{Word} & No-Drift   & Additive & TransDrift \\ \midrule
\textit{well}                   & 7    & 4 & \textbf{7}     \\ 
\textit{place}                       & 8 & 8       & \textbf{15}       \\ 
\textit{great}                           & 10  & 11     & \textbf{13}      \\ 
\textit{time}                         & 7   & 7  & \textbf{12}     \\ 
\textit{nice} & 11 & 9 & \textbf{15} \\
\textit{customer} & 8 & 9 & \textbf{10} \\
\textit{happy} & 3 & 1 & \textbf{7} \\
\textit{people} & 9 & 8 & \textbf{12} \\
\bottomrule
\end{tabular}
\caption{\textbf{Qualitative analysis of nearest neighbors of the predicted word embeddings.} For each prediction model, we find 30 nearest neighbors for each word shown in the first column. We then count the number of these nearest neighbors that are also the nearest neighbor in the target word embeddings. Thus, the higher number of nearest neighbors of our model TransDrift shows that our predicted embeddings agree significantly more with the target embeddings.}
\label{tab:qual-nn-count}
\end{table}

% {\color{red}\lipsum[1]}

\subsection{Downstream Tasks}

We now evaluate how well our predicted embeddings can enable better performance in downstream tasks. In particular, we seek to evaluate that as the data undergoes drift, can embedding prediction help the accuracy under the drifted distribution (at time-step 2) and if so then which prediction approach should be preferred.
% Formally, can prediction help downstream accuracy for the data $\cD_2$ and if so then which prediction approach should be preferred?
We consider the following approaches for obtaining word embedding under drift and compare with our model: \textit{i)} \textit{No-Drift model}: We compare with this approach as in deployment setting, this is often considered as the default approach. \textit{ii)} \textit{Full Retraining.} Another common approach is to retrain the embedding from scratch from the drifted distribution. However, note that in some cases, this can be an unfair comparison to our model as the data from the drifted distribution may be either too little or not available. For our model TransDrift, we also leverage embeddings from $\cD_2^\text{small}$ having 30\% of the data of full $\cD_2$.

We train a downstream classification model using the embedding from each of the evaluated methods and report the test accuracy in Table \ref{tab:downstream-quant-1} for the Amazon Review dataset. We find that the No-Drift model which re-uses the outdated embedding from the previous time-step suffers with respect to our model. This suggests that embedding prediction is indeed useful. 
%Furthermore, the embedding $\bE_2$ trained on a large dataset taken from the drifted distribution also suffers. 
We further analyze the downstream performance by showing qualitative examples of text inputs from the drifted distribution that were misclassified by the No-Drift model but are correctly classified by our model TransDrift. We show these in Table \ref{tab:downstream-qual}.

\subsection{Ablation Study}
In this section, we ablate our model TransDrift. To better justify our choice of architecture for TransDrift, we perform additional experiments that we describe here. In terms of architectural components our model can be seen as Self-Attention + Feed Forward Network while our baseline MLP can be seen as Feed Forward Network. We analyze the effect of this choice in our experiments. We provide the results in Appendix \ref{sec:appendix}. The results show that the TransDrift outperforms the ablation model. This shows that the self-attention aspect of TransDrift plays a crucial role and not just MLP.

\begin{table}[t]
\centering
\scalebox{0.85}{
\begin{tabular}{@{}lccc@{}}
\toprule
& \multicolumn{3}{c}{Accuracy (in \%)} \\ \cmidrule(l){2-4}
{Dataset}       & No-Drift  & Additive    &   {TransDrift}     \\ \midrule
AR-Electro    & 60.5\%  & 60.56\%  & \textbf{69.6\%}  \\ 
AR-Kitchen        & 63.6\%   &  63.52\%   & \textbf{75.7\%}    \\ 
AR-DVD            & 59.0\%    &   59.03\%  & \textbf{63.5\%}  \\
Yelp           & 58\%     & 60\% & \textbf{65.0\%}  \\
\bottomrule
\end{tabular}}
\caption{\textbf{Downstream Prediction Results on Amazon Review (AR) and Yelp Datasets.} Using embeddings from the evaluated methods, we train a downstream sentiment classifier and report its test accuracy. We note that the No-Drift model which re-uses the outdated embedding from the previous time-step suffers compared to TransDrift. TransDrift is significantly more accurate than the baseline models.}
\label{tab:downstream-quant-1}
\end{table}

%backup downstream
% \begin{table}[t]
% \centering
% \scalebox{0.85}{
% \begin{tabular}{@{}lccc|c@{}}
% \toprule
% & \multicolumn{4}{c}{Accuracy (in \%)} \\ \cmidrule(l){2-5}
% {Dataset}       & No-Drift  & Additive    &   {TransDrift}  & $\bE_2$     \\ \midrule
% AR-Electro    & 60.5\%  & 60.56\%  & \textbf{69.6\%}  & 61.4\% \\ 
% AR-Kitchen        & 63.6\%   &  63.52\%   & \textbf{75.7\%} & 62.3\%    \\ 
% AR-DVD            & 59.0\%    &   59.03\%  & \textbf{63.5\%}  & 60.8\%  \\
% Yelp           & 58\%     & 60\% & \textbf{65.0\%}  & {69\%} \\
% \bottomrule
% \end{tabular}}
% \caption{\textbf{Downstream Prediction Results on Amazon Review (AR) and Yelp Datasets.} Using embeddings from the evaluated methods, we train a downstream sentiment classifier and report its test accuracy. We note that the No-Drift model which re-uses the outdated embedding from the previous time-step suffers compared to TransDrift. TransDrift is significantly more accurate than the baseline models. Furthermore, we report the accuracy with the embedding $\bE_2$ trained on a large dataset from the drifted distribution. This can be expected to act as an upperbound since these embeddings conform to the drifted distribution. However,  surprisingly, our model TransDrift outperforms this in Amazon Review dataset.}
% \label{tab:downstream-quant-1}
% \end{table}

\begin{table*}[t]
\centering
\small
\begin{tabular}{@{}lcc@{}}
\toprule
\multicolumn{1}{c}{\textbf{Review Text}}                                                                                                                                                                                                                                                                                                                                                                                                                                                                                                                                                                                                                                                                                                                                                                                                      & \textbf{Ground Truth} & \textbf{TransDrift} \\ \midrule
\begin{tabular}[c]{@{}l@{}}First off, the ipod jiggles no matter what you do, secondly, it doesn't stay straight \\ on the power plug, it constantly tilts(the whole thing)...not worth \$10\end{tabular}                                                                                                                                                                                                                                                                                                                                                                                                                                                                                                                                                                                                                            & Negative     & Negative  \\ \midrule
\begin{tabular}[c]{@{}l@{}}I bought this amazing product and now it is easy to have high quality music. \\ Just plug the iPod to your music equipment and you are done.\end{tabular}                                                                                                                                                                                                                                                                                                                                                                                                                                                                                                                                                                                                                                                 & Positive     & Positive  \\ \midrule

% \begin{tabular}[c]{@{}l@{}}I would like to believe that this hub works well- but I wouldn't know. I bought it\\  to reduce clutter on my desk from all my devices hooking into my powermac. \\ I hooked up my printer, ipod and digital camera and nothing happened. \\ I called Targus and they said I needed an a/c adaptor. It doesn't come with one\\ and doesn't say anything about one in the manual or instructions, or their \\ website. They said I could get it at radio shack and gave me the info \\ (5 volt, 2.1 amp, center positive). I've been to two radio \\ shacks and online- they don't carry it. I'm upset. Any advice\end{tabular}                                                                                                                                                                            & Negative     & Negative  \\ \midrule

\begin{tabular}[c]{@{}l@{}}I just got my mouse today I was ecstatic about the performance I had initial\\  problems installing the mouse but after I unplugged it and plugged it in again\\  the problems went away. I have to agree with my other comrades from this site \\ this is the best mouse for the price and I have no problems whatsoever \\ except for the look which I thought was the result from damage from shipping\\  but it was designed like that so I have no complaints
\end{tabular}                                                                                                                                                                                                                                                                                                                           & Positive     & Positive  \\ \bottomrule
\end{tabular}
\caption{Text samples that were misclassified when using No-Drift model compared to our TransDrift model.}
\label{tab:downstream-qual}
\end{table*}

\section{Related Work}
% {\color{red}\lipsum[1]}

% \textbf{Meta-Learning.}

\textbf{Word Vectors.} Learning word representations has seen significant interest in the past decade \citep{word2vec2013, bojanowski2017enriching, barkan2021representation, bojanowski2017fastText, faruqui2015retrofitting, caciularu2021denoising, bollegala2016joint, hu2019few}. The most common approach has been proposed by \cite{word2vec2013} providing two architectures, CBOW and skip-gram, for learning high-quality word vectors from large text datasets. CBOW learns by predicting the current word based on the context words, whereas skip-gram predicts the nearby context words. \citep{bojanowski2017fastText} discover a new method for learning word representation which incorporated character $n$-gram to the skipgram model. They ensure that the model takes sub-word information into account, improving embedding quality, and predicting the embeddings for unseen words.

% Semantic and syntactic relationships are tested to measure the quality of word vectors.  --
%   ----

\textbf{Data Drift in Text.} While analyzing the presence of drift in text has seen significant interest in recent years, efforts to model their drift are still in infancy. \cite{huang2018examining} examine the problem of drift and that show its adverse effects on the downstream performance if training and test sets are not the same distributions due to drift. \citep{leszczynski2020understanding, Chugh2018StabilityOW} define task instability with respect to word embedding (and the task being done) and propose a metric to measure it. Using this metric, trade-off in stability with respect to precision and model dimension are identified. \cite{hellrich2016bad} and \citep{antoniak-mimno-2018-evaluating} identify instability in word neighbors between different training runs in \textit{word2vec} and \textit{fasttext} embeddings. \citep{wendlandt2018factors, pierrejean2018predicting} define stability as percent overlap among neighbors which, crucially, serves as a task independent definition. Analysis of various factors that affect word stability and their effects on downstream tasks was also performed. To address embedding instability, \citep{hellrich2018influence} propose a down-sampling based approach to make word embedding more stable. \citep{stowe2021combating} propose a reversal of drift in word embedding to make it stable over time while \cite{he2018time} propose an evolutionary approach. However, all these approaches only focus on making embedding more stable under the assumption that the downstream task should remain agnostic to the drift in the underlying data distribution. In contrast, our work seeks to improve the embedding and the task performance by taking drift and changing word semantics into account. \cite{xu2018lifelong} propose meta-learning approach to adapt word embedding from source to target domain. However unlike our method, this approach requires direct access to all the corpora of the previously seen domains and is thus orthogonal to our problem setting. Unlike our method, this approach cannot be applied if there is no available data from the target domain.\\

\textbf{Contextual Embeddings.} Contextual embeddings have also seen a rise alongside word2vec. However, word2vec is widely used in a lot of industrial applications  \citep{twitterfeature, tensorhub, feast, derczynski2015usfd, fromreide2014crowdsourcing}, the scope of our work is to deal with drift in regular word2vec embeddings. 
 
\section{Conclusion}
% {\color{red}\lipsum[3]}
In this paper, we proposed TransDrift, a framework to track embeddings under data drift. We showed that using a transformer model perform this task effectively with no data. Optionally, our model can also leverage small amount of data from drifted distribution to further improve its prediction. Finally, by performing downstream tasks using the predicted embeddings, we show a significant performance improvement compared to other options. 
One of the future work can be to study multi-step word embedding prediction.
% \bibliography{custom}

% \subsection{References}

% \nocite{Ando2005,borschinger-johnson-2011-particle,andrew2007scalable,rasooli-tetrault-2015,goodman-etal-2016-noise,harper-2014-learning}

% The \LaTeX{} and Bib\TeX{} style files provided roughly follow the American Psychological Association format.
% If your own bib file is named \texttt{custom.bib}, then placing the following before any appendices in your \LaTeX{} file will generate the references section for you:
% \begin{quote}
% \begin{verbatim}
\bibliographystyle{acl_natbib}
\bibliography{custom}
% \end{verbatim}
% \end{quote}

% \bibliographystyle{acl_natbib}

\clearpage
\appendix
\section{Additional Experiment Details}
\label{sec:appendix}
\subsection{Ablation Study}
To better justify our choice of architecture for TransDrift, we perform additional experiments that we describe here. In terms of architectural components our model can be seen as Self-Attention + Feed Forward Network while our baseline MLP can be seen as Feed Forward Network. We analyze the effect of this choice in our experiments. We shown in Table \ref{tab:quant-cosine-baseline} the cosine similarity of the predicted embedding with the ground truth embedding trained using large amount of data from the drifted distribution. We note that our model, TransDrift, is significantly more accurate with respect to the MLP model. Using embeddings from the evaluated methods, we train a downstream sentiment classifier and report its test accuracy in Table \ref{tab:downstream-quant}. We show a comparison between MLP and Transdrift. We note that the MLP model suffers compared to TransDrift. TransDrift is significantly more accurate than the baseline models.

\begin{table}[h!]
\centering

\begin{tabular}{llc}
\toprule
\textbf{} & \multicolumn{2}{c}{Drift Model}                \\ \cmidrule(l){2-3} 
Dataset   &  MLP    & TransDrift      \\ \midrule
Synthetic &     \textbf{0.785}    & 0.773 \\
Yelp      &   0.89     & \textbf{0.89} \\
Amazon    & 0.231 & \textbf{0.8170} \\ 
\bottomrule
\end{tabular}

% \end{table}

% \begin{table}[t]
% \centering
% \begin{tabular}{@{}lccc@{}}
% \toprule
% & \multicolumn{3}{c}{Drift Model} \\\cmidrule(l){2-4}
%   Dataset            & {No-Drift} & {Additive} &
%   {TransDrift} \\ \midrule
% Synthetic            & 0.32     & 0.33      &     \textbf{0.7724}          \\ 
% Yelp & 0.19 & 0.7956  &  \textbf{0.8910} \\ 
% Amazon             & -0.004   & -0.0002          &  \textbf{0.8170}             \\ \bottomrule
% \end{tabular}
\caption{\textbf{Comparison of word embedding prediction between our model and the baselines.} We report the cosine similarity of the predicted embedding with the ground truth embedding trained using large amount of data from the drifted distribution. The predicted embeddings do not use any data from the drifted distribution. We note that our model, TransDrift, is significantly more accurate with respect to the MLP model.}
\label{tab:quant-cosine-baseline}
\end{table}

\begin{table}[h!]
\centering
\scalebox{0.85}{
\begin{tabular}{@{}lcc@{}}
\toprule
& \multicolumn{2}{c}{Accuracy (in \%)} \\ \cmidrule(l){2-3}
{Dataset}       & MLP   & {TransDrift}    \\ \midrule
AR-Electro    & 60.81\% & \textbf{69.6\%}   \\ 
AR-Kitchen        & 55.72\% & \textbf{75.7\%}    \\ 
AR-DVD            &  55.01\% & \textbf{63.5\%}  \\
Yelp           & 59\% & \textbf{65.0\%}  \\
\bottomrule
\end{tabular}}
\caption{\textbf{Downstream Prediction Results on Amazon Review (AR) and Yelp Datasets.} Using embeddings from the evaluated methods, we train a downstream sentiment classifier and report its test accuracy. We show a comparison between MLP and Transdrift. We note that the MLP model suffers compared to TransDrift. TransDrift is significantly more accurate than the baseline models.}
\label{tab:downstream-quant}
\end{table}

\clearpage

\subsection{Model Hyperparameters}

\textbf{MLP}

\begin{itemize}
\item Model parameters = 20250
\end{itemize}
\begin{itemize}
\item Max-epochs = 50, model-dim = 50, warmup = 30, LR = 5e-4, Batch size = 100
\end{itemize}

\section{Synthetic Dataset}

\begin{itemize}
\item Common words for the 1000 set of embeddings of :-
\begin{enumerate}
\item D1 = 100,
\item D2 = 100,
\item D2-small-50 = 50,
\item D2-small-30 = 30,
\item D2-small-20 = 20

\end{enumerate}
\end{itemize}
\begin{itemize}
\item Dimension for D1, D2, D2-small-s and predicted embedding is 50.
\end{itemize}

\textbf{TransDrift}

\begin{itemize}
\item Model parameters - 504250
\end{itemize}
\begin{itemize}
\item Max-epochs = 100, num-heads = 1, model-dim = 100, num-layers = 4, LR = 5e-4, Batch size = 100
\end{itemize}

\section{Yelp Dataset}

\begin{itemize}
\item Common words for the 1000 set of embeddings of :-
\begin{enumerate}
\item D1 = 1727,
\item D2 = 1727,
\item D2-small-50 = 1396,
\item D2-small-30 = 901,
\item D2-small-20 = 634
\end{enumerate}
\end{itemize}
            
\begin{itemize}
\item Dimension for D1, D2, D2-small-s and predicted embedding is 50.
\end{itemize}
 
 \textbf{TransDrift}
Ran on CPU.
\begin{itemize}
\item Model parameters - 1833650
\end{itemize}

\begin{itemize}
\item Using word2vec, Max-epochs = 100, num-heads = 4, model-dim = 192, num-layers = 4, LR = 5e-4, Batch size = 100
\end{itemize}

\section{Amazon Customer Review Dataset}

\begin{itemize}
\item Common words for the 1000 set of embeddings of :-
\begin{enumerate}
\item D1 = 5018,
\item D2 = 5018,
\item D2-small-50 = 3323,
\item D2-small-30 = 2235,
\item D2-small-20 = 1603
\end{enumerate}
\end{itemize}
    
\begin{itemize}
\item Dimension for D1, D2, D2-small-s and predicted embedding is 50.
\end{itemize}

\textbf{TransDrift}
Ran on CPU.
\begin{itemize}
\item Model paramters - 1833650
\end{itemize}
\begin{itemize}
\item Using word2vec, Max-epochs = 100, num-heads = 4, model-dim = 192, warmup = 30, num-layers = 4, LR = 5e-4, Batch size = 100
\end{itemize}

\section{Downstream}
\begin{itemize}
\item E1 and $E2\_small\_30$ are used to predict the embeddings using transformer. These predicted embeddings are used in the downstream tasks.
\end{itemize}
 
\subsection{Synthetic Dataset}
Not done

\subsection{Yelp Dataset}

Multi-label star classification

\begin{itemize}
\item 20k reviews : 10k from each time stamp t1 and t2
\item 8:2 train-test split, epochs = 100, batch-size = 64
\end{itemize}

% Model Summary

Model: "model"                                                                                                               
\begin{table*}[h!]
\centering
\begin{tabular}{@{}lll@{}}
\toprule
\textbf{Layer (type)} & \textbf{Output Shape} & \textbf{Param \#} \\ \midrule
input\_1 (InputLayer) & (None, 150)           & 0                 \\
embedding (Embedding) & (None, 150, 50)       & 1611750           \\
lstm (LSTM)           & (None, 128)           & 91648             \\
dense (Dense)         & (None, 5)             & 645               \\ \bottomrule
\end{tabular}
\caption{\textbf{Model architecture for multi-label classification task.} We present the layers contained in the LSTM-based model used to do classification on the yelp dataset on star (rating).}

\end{table*}

Total params: 1,704,043
Trainable params: 92,293
Non-trainable params: 1,611,750

\subsection{Amazon Customer Review Dataset}

\begin{itemize}
\item Performed sentiment analysis task on electronics, kitchen, and dvd reviews
\item Number of reviews used for sentiment analysis task:-
\begin{enumerate}
\item  Electronics - 5682
\item Kitchen - 5946
\item DVD - 3587
\end{enumerate}
\end{itemize}

\begin{itemize}
\item 8:2 train-test split, epochs = 100, batch-size = 64
\end{itemize}

%Model summary

Model: "model"               

% \usepackage{booktabs}
% \usepackage{booktabs}
\begin{table*}[h!]
\centering

\begin{tabular}{@{}lll@{}}
\toprule
\textbf{Layer (type)} & \textbf{Output Shape} & \textbf{Param \#} \\ \midrule
input\_1 (InputLayer) & (None, 150)           & 0                 \\
embedding (Embedding) & (None, 150, 50)       & 747250            \\
lstm (LSTM)           & (None, 150, 128)      & 91648             \\
dropout (Dropout)     & (None, 150, 128)      & 0                 \\
lstm\_1 (LSTM)        & (None, 150, 128)      & 131584            \\
dropout\_1 (Dropout)  & (None, 150, 128)      & 0                 \\
lstm\_2 (LSTM)        & (None, 128)           & 131584            \\
dense (Dense)         & (None, 1)             & 129               \\ \bottomrule
\end{tabular}
\caption{\textbf{Model architecture for Sentiment Analysis task.} We report the layers present in the LSTM based model used to perform sentiment analysis on the Amazon dataset.}
\end{table*}

Total params: 1,102,195
Trainable params: 354,945
Non-trainable params: 747,250

\subsection{Qualitative Results}
We show the qualitative samples in Table \ref{tab:downstream-quals} that were misclassified when using No-Drift model compared to our TransDrift model.

\begin{table*}[t]
\centering
\small
\begin{tabular}{@{}lcc@{}}
\toprule
\multicolumn{1}{c}{\textbf{Review Text}}                                                                                                                                                                                                                                                                                                                                                                                                                                                                                                                                                                                                                                                                                                                                                                                                      & \textbf{Ground Truth} & \textbf{TransDrift} \\ \midrule
\begin{tabular}[c]{@{}l@{}}First off, the ipod jiggles no matter what you do, secondly, it doesn't stay straight \\ on the power plug, it constantly tilts(the whole thing)...not worth \$10\end{tabular}                                                                                                                                                                                                                                                                                                                                                                                                                                                                                                                                                                                                                            & Negative     & Negative  \\ \midrule
\begin{tabular}[c]{@{}l@{}}I bought this amazing product and now it is easy to have high quality music. \\ Just plug the iPod to your music equipment and you are done.\end{tabular}                                                                                                                                                                                                                                                                                                                                                                                                                                                                                                                                                                                                                                                 & Positive     & Positive  \\ \midrule
\begin{tabular}[c]{@{}l@{}}This printer was very easy to use. Just pop in the cartridge, paper, and plug in \\ the camera. The camera has to be a certain type of Canon, but there is\\  an available computer cable (I have not gotten it yet), to hook up other cameras.\\  The benefit of using a Canon camera with this printer is you can travel\\  without a computer, and still print out pictures. These pictures look beautiful,\\  and the colors are true and bright. Printing couldn't be easier and fast. I \\ was able to get a print while my 2 year old was frantically trying\\  to pull the plugs, press the buttons  and all arms and legs struggling. \\ It's fun watching the three colors slide in and out in the three \\ passes the printer makes to process the picture. Instant gratification\end{tabular} & Positive     & Positive  \\ \midrule

% \begin{tabular}[c]{@{}l@{}}I would like to believe that this hub works well- but I wouldn't know. I bought it\\  to reduce clutter on my desk from all my devices hooking into my powermac. \\ I hooked up my printer, ipod and digital camera and nothing happened. \\ I called Targus and they said I needed an a/c adaptor. It doesn't come with one\\ and doesn't say anything about one in the manual or instructions, or their \\ website. They said I could get it at radio shack and gave me the info \\ (5 volt, 2.1 amp, center positive). I've been to two radio \\ shacks and online- they don't carry it. I'm upset. Any advice\end{tabular}                                                                                                                                                                            & Negative     & Negative  \\ \midrule

\begin{tabular}[c]{@{}l@{}}I just got my mouse today I was ecstatic about the performance I had initial\\  problems installing the mouse but after I unplugged it and plugged it in again\\  the problems went away. I have to agree with my other comrades from this site \\ this is the best mouse for the price and I have no problems whatsoever \\ except for the look which I thought was the result from damage from shipping\\  but it was designed like that so I have no complaints
\end{tabular}                                                                                                                                                                                                                                                                                                                           & Positive     & Positive  \\ \bottomrule
\end{tabular}
\caption{Text samples that were misclassified when using No-Drift model compared to our TransDrift model.}
\label{tab:downstream-quals}
\end{table*}

\end{document}